\documentclass[10pt,english,british,10pt,twocolumn,letterpaper]{article}
\usepackage[T1]{fontenc}
\usepackage[latin9]{inputenc}
\usepackage{array}
\usepackage{units}
\usepackage{enumitem}
\usepackage{multirow}
\usepackage{amsmath}
\usepackage{graphicx}

\makeatletter

\providecommand{\tabularnewline}{\\}
\newcommand{\lyxdot}{.}

\usepackage{cvpr}
\usepackage{times}
\usepackage{epsfig}
\usepackage{graphicx}
\usepackage{amsmath}
\usepackage{amssymb}

\usepackage{placeins}
\usepackage{color}
\usepackage{colortbl}
\usepackage{rotating}
\definecolor{lightgray}{gray}{0.8}
\definecolor{verylightgray}{gray}{0.9}

\usepackage[pagebackref=true,breaklinks=true,letterpaper=true,colorlinks,bookmarks=false]{hyperref}

 \cvprfinalcopy 

\def\cvprPaperID{1191} 

\ifcvprfinal\fi

\@ifundefined{showcaptionsetup}{}{%
 \PassOptionsToPackage{caption=false}{subfig}}
\usepackage{subfig}
\makeatother

\usepackage{babel}
\begin{document}

\let\originalparagraph\paragraph 
\renewcommand{\paragraph}[2][.]{\originalparagraph{#2#1}}
\setlist[enumerate]{itemindent=\dimexpr\labelwidth+\labelsep\relax,leftmargin=0pt}
\setlist[description]{itemindent=\dimexpr\labelwidth+\labelsep\relax,leftmargin=0pt}
\setlength{\parskip}{0pt}

\title{CityPersons: A Diverse Dataset for Pedestrian Detection}

\author{Shanshan Zhang, Rodrigo Benenson and Bernt Schiele\\
\begin{tabular}{c}
Max Planck Institute for Informatics\tabularnewline
Saarbrücken, Germany\tabularnewline
\texttt{\small{}firstname.lastname@mpi-inf.mpg.de}\tabularnewline
\end{tabular}\vspace{-1em}
 \and}
\maketitle
\begin{abstract}
Convnets have enabled significant progress in pedestrian detection
recently, but there are still open questions regarding suitable architectures
and training data. We revisit CNN design and point out key adaptations,
enabling plain FasterRCNN to obtain state-of-the-art results on the
Caltech dataset. 

To achieve further improvement from more and better data, we introduce
\texttt{CityPersons}, a new set of person annotations on top of the
Cityscapes dataset. The diversity of \texttt{CityPersons} allows us
for the first time to train one single CNN model that generalizes
well over multiple benchmarks. Moreover, with additional training
with \texttt{CityPersons}, we obtain top results using FasterRCNN
on Caltech, improving especially for more difficult cases (heavy occlusion
and small scale) and providing higher localization quality.
\end{abstract}

\section{\label{sec:Introduction}Introduction}

Pedestrian detection is a popular topic in computer vision community,
with wide applications in surveillance, driving assistance, mobile
robotics, etc. During the last decade, several benchmarks have been
created for this task \cite{Dalal2005Cvpr,Dollar2014Pami,Geiger2012CVPR}.
These benchmarks have enabled great progress in this area \cite{Benenson2014Eccvw}. 

While existing benchmarks have enabled progress, it is unclear how
well this progress translate in open world performance. We think it
is time to give emphasis not only to intra-dataset performance, but
also across-datasets.

Lately, a wave of convolutional neural network (convnet) variants
have taken the Caltech benchmark top ranks \cite{Zhang2016Eccv,cai16mscnn,SA-FastRCNN,cai2015iccv,tian2015iccv}.
Many of these are custom architectures derived from the FasterRCNN
\cite{Girshick2014Cvpr,FastRCNN,renNIPS15fasterrcnn} general object
detector. We show here that a properly adapted FasterRCNN can match
the detection quality of such custom architectures. However since
convnets are high capacity models, it is unclear if such model will
benefit from more data.

To move forward the field of pedestrian detection, we introduce ``\texttt{\small{}CityPersons}'',
a new set of annotations on top of Cityscapes \cite{Cordts2016Cityscapes}.
These are high quality annotations, that provide a rich diverse dataset,
and enable new experiments both for training better models, and as
new test benchmark.

In summary, our main contributions are:
\begin{figure}
\begin{centering}
\includegraphics[width=0.45\textwidth]{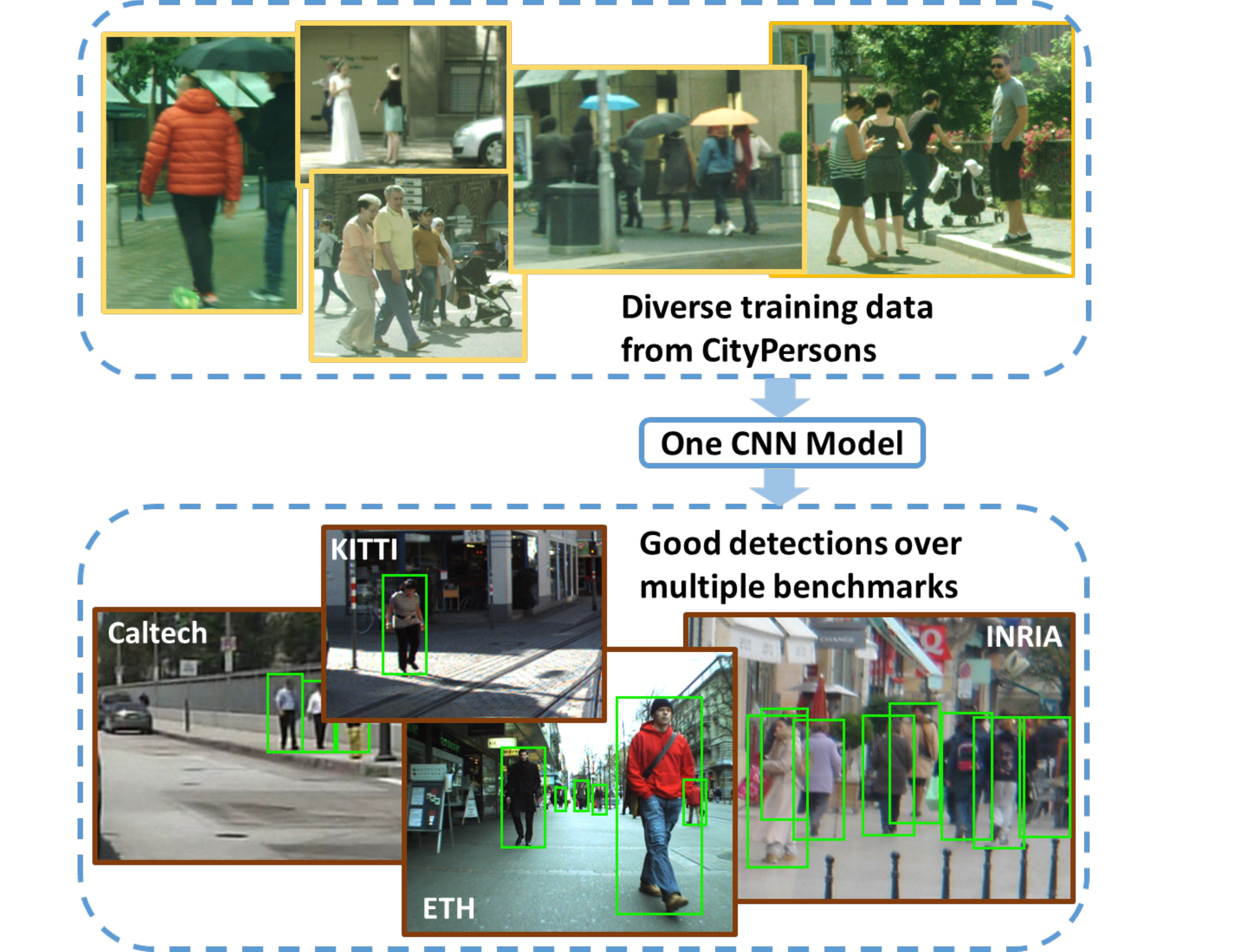}
\par\end{centering}
\caption{The diversity of the newly introduced CityPersons annotations allows
to train one convnet model that generalizes well over multiple benchmarks.}
\end{figure}
\vspace{-0.5em}

\begin{enumerate}
\item We introduce CityPersons, a new set of high quality bounding box annotations
for pedestrian detection on the Cityscapes dataset (train, validation,
and test sets). The train/val. annotations will be public, and an
online benchmark will be setup.\vspace{-0.5em}
\item We report new state-of-art results for FasterRCNN on Caltech and KITTI
dataset, thanks to properly adapting the model for pedestrian detection
and using \texttt{\small{}CityPersons} pre-training. We show in particular
improved results for more difficult detection cases (small and occluded),
and overall higher localization precision. \vspace{-0.5em}
\item Using CityPersons, we obtain the best reported across-dataset generalization
results for pedestrian detection.\vspace{-0.5em}
\item We show preliminary results exploiting the additional Cityscapes annotations.
Using semantic labelling as additional supervision, we obtain promising
improvements for detecting small persons.\vspace{-0.5em}
\end{enumerate}
Section \ref{subsec:Related-work} covers the related work, section
\ref{sec:pedestrian-convnet} discusses how to adapt FasterRCNN for
best detection quality, section \ref{sec:CityPersons} describes our
annotation process, some statistics of the new data and baseline experiments.
Finally, section \ref{sec:Improve-quality} explores different ways
to use \texttt{\small{}CityPersons} to improve person detection quality.

\subsection{\label{subsec:Related-work}Related work}

In this paper, we investigate convnets, datasets and semantic labels
for pedestrian detection, so we discuss related works for these three
aspects.

\paragraph{Convnets for pedestrian detection}

Convolutional neural networks (convnets) have achieved great success
in classification and detection on the ImageNet \cite{Krizhevsky2012Nips},
Pascal, and MS COCO datasets \cite{He2016Cvpr}. FasterRCNN \cite{Girshick2014Cvpr,FastRCNN,renNIPS15fasterrcnn}
has become the de-facto standard detector architecture. Many variants
work try to extend it \cite{YOLO,cai16mscnn,densebox}, but few improve
results with a simpler architecture. A notable exception is SSD \cite{liu15ssd},
which obtains comparable results with a simpler architecture.

Initial attempts to apply convnets for pedestrian detection, used
existing detectors (mainly decision forests over hand-crafted features
\cite{Benenson2014Eccvw,Zhang2015Cvpr}) outputs and re-scored them
with a convnet classifier (plus bounding box regression) \cite{Hosang2015Cvpr,Yang2015Cvpr,Angelova2015Bmvc,tian2015iccv}.
Better results are shown when using the reverse configuration: detections
resulted from a convnet are re-scored with decision forests classifier
(trained over convnet features) \cite{cai2015iccv,Zhang2016Eccv}.
Recently good results are presented by customized pure convnet architectures
such as MS-CNN \cite{cai16mscnn} and SA-FastRCNN \cite{SA-FastRCNN}.\\
In this paper we show that a properly adapted plain FasterRCNN matches
state-of-the-art detection quality without needing add-ons.

\paragraph{Pedestrian datasets}

In the last decade several datasets have been created for pedestrian
detection training and evaluation. INRIA \cite{Dalal2005Cvpr}, ETH
\cite{Ess2008Cvpr}, TudBrussels \cite{Wojek2009Cvpr}, and Daimler
\cite{Enzweiler2009PAMI} represent early efforts to collect pedestrian
datasets. These datasets have been superseded by larger and richer
datasets such as the popular Caltech-USA \cite{Dollar2012Pami} and
KITTI \cite{Geiger2012CVPR}. Both datasets were recorded by driving
through large cities and provide annotated frames on video sequences.\\
Despite the large number of frames, both datasets suffer from low-density.
With an average of $\sim\negmedspace1$ person per image, occlusions
cases are severely under-represented. Another weakness of both dataset,
is that each was recorded in a single city. Thus the diversity in
pedestrian and background appearances is limited.\\
Building upon the strengths of the Cityscapes data \cite{Cordts2016Cityscapes},
our new annotations provide high quality bounding boxes, with larger
portions of occluded persons, and the diversity of 27 different cities.
Such diversity enables models trained on \texttt{\small{}CityPersons}
to better generalize to other test sets.

\paragraph{Semantic labels for pedestrian detection}

In section \ref{subsec:Semantic-labels} we will explore using the
semantic labels from Cityscapes to train a pedestrian detector with
better context modelling. The idea of using semantic labels to improve
detections is at least a decade old \cite{Wolf2006Ijcv}, and two
recent incarnations are \cite{Hu2016arxiv,CosteaCVPR16}. We will
use the semantic probability maps computed from a semantic labeller
network as additional input channels (next to RGB channels) for the
pedestrian detection convnet (see section \ref{subsec:Semantic-labels}).

\section{\label{sec:pedestrian-convnet}A convnet for pedestrian detection}

Before delving into our new annotations (in section \ref{sec:CityPersons}),
we first build a strong reference detector, as a tool for our experiments
in sections \ref{sec:Baseline-experiments} and \ref{sec:Improve-quality}.
We aim at finding a straightforward architecture that provides good
performance on the Caltech-USA dataset \cite{Dollar2012Pami}.

\paragraph{Training, testing ($\mbox{MR}^{O}$, $\mbox{MR}^{N}$)}

We train our Caltech models using the improved $10\times$ annotations
from \cite{shanshan_cvpr16}, which are of higher quality than the
original annotations (less false positives, higher recall, improved
ignore regions, and better aligned bounding boxes). For evaluation
we follow the standard Caltech evaluation \cite{Dollar2012Pami};
log miss-rate (MR) is averaged over the FPPI (false positives per
image) range of $[10^{-2},\,10^{0}]\ \text{FPPI}$. Following \cite{shanshan_cvpr16},
we evaluate both on the ``original annotations'' ($\mbox{MR}^{O}$)
and new annotations ($\mbox{MR}^{N}$); and indicate specifically
which test set is being used each time. Unless otherwise specified,
the evaluation is done on the ``reasonable'' setup \cite{Dollar2012Pami}.

\paragraph{\label{subsec:faster-rcnn}FasterRCNN}

The FasterRCNN detector obtains competitive performance on general
object detection. After re-training with default parameters it will
under-perform on the pedestrian detection task (as reported in \cite{Zhang2016Eccv}).
The reason why vanilla FasterRCNN underperforms on the Caltech dataset
is that it fails to handle small scale objects ($50\negmedspace\sim\negmedspace70\ \text{pixels}$),
which are dominant on this dataset. To better handle small persons,
we propose five modifications ($\text{M}_{i}$) that bring the $\text{MR}^{O}$
(miss-rate) from $20.98$ down to $10.27$ (lower is better, see table
\ref{tab:improvements-step-by-step}). As of writing, the best reported
results on Caltech is $9.6\ \text{MR}^{o}$, and our plain FasterRCNN
ranks third with less than a point difference. We train FasterRCNN
with VGG16 convolutional layers, initialized via ImageNet classification
pre-training \cite{renNIPS15fasterrcnn}.
\begin{table}
\begin{centering}
\setlength{\tabcolsep}{4pt} 
\begin{tabular}{l|c|c}
Detector aspect & $\mbox{MR}^{O}$ & $\Delta\mbox{MR}$\tabularnewline
\hline 
\hline 
\texttt{FasterRCNN-vanilla} & 20.98 & -\tabularnewline
+ quantized rpn scales & 18.11 & + 2.87\tabularnewline
+ input up-scaling  & 14.37 & + 3.74\tabularnewline
+ Adam solver & 12.70 & + 1.67\tabularnewline
+ ignore region handling & 11.37 & + 1.33\tabularnewline
+ finer feature stride & 10.27 & + 1.10\tabularnewline
\hline 
\texttt{FasterRCNN-ours} & 10.27 & + 10.71\tabularnewline
\end{tabular}
\par\end{centering}
\caption{\label{tab:improvements-step-by-step}Step by step improvements on
Caltech from vanilla FasterRCNN to our adapted version, we gain $10.71\ \text{MR}$
points in total.}
\vspace{-0.5em}
\end{table}

\begin{description}
\item [{$\text{M}_{1}$~Quantized~RPN~scales.}] The default scales of
the RPN (region proposal network in FasterRCNN) are sparse ($[0.5,\,1,\,2]$)
and assume a uniform distribution of object scales. However, when
we look at the training data on Caltech, we find much more small scale
people than large ones. Our intuition is to let the network generate
more proposals for small sizes, so as to better handle them. We split
the full scale range in $10$ quantile bins (equal amount of samples
per bin), and use the resulting $11$ endpoints as RPN scales to generate
proposals.
\item [{$\text{M}_{2}$~Input~up-scaling.}] Simply up-sampling the input
images by 2x, provides a significant gain of $3.74\ \text{MR}^{O}$
percent points (pp). We attribute this to a better match with the
ImageNet pre-training appearance distribution. Using larger up-sampling
factors does not show further improvement. 
\item [{$\text{M}_{3}$~Finer~feature~stride.}] Most pedestrians in
Caltech have $\text{height}\times\text{width}=80\times40$. The default
VGG16 has a feature stride of 16 pixels. Having such a coarse stride
compared to the object width reduces the chances of having a high
score over persons, and forces the network to handle large displacement
relative to the object appearance. Removing the fourth max-pooling
layer from VGG16 reduces the stride to 8 pixels; helping the detector
to handle small objects.
\item [{$\text{M}_{4}$~Ignore~region~handling.}] The vanilla FasterRCNN
code does not cope with ignore regions (areas where the annotator
cannot tell if a person is present or absent, and person groups where
individuals cannot be told apart). Simply treating these regions as
background introduces confusing samples, and has a negative impact
on the detector quality. By ensuring that during training the RPN
proposals avoid sampling the ignore regions, we observe a $1.33\ \text{MR\,\ensuremath{\text{pp}}}$
improvement.
\item [{$\text{M}_{5}$~Solver.}] Switching from the standard Caffe SGD
solver to the Adam solver \cite{adamICLR15}, provides a consistent
gain in our experiments.
\end{description}
We show the step-by-step improvements in table \ref{tab:improvements-step-by-step}.
$\text{M}_{1}$ and $\text{M}_{2}$ are key, while each of the other
modifications add about $\sim\negmedspace1\ \text{MR\,\ensuremath{\text{pp}}}$.
All together these modifications adapt the vanilla FasterRCNN to the
task of pedestrian detection.

\paragraph{\label{subsec:other-architectures}Other architectures}

We also explored other architectures such as SSD \cite{liu15ssd}
or MS-CNN \cite{cai16mscnn} but, even after adaptations, we did not
manage to obtain improved results. Amongst all the variants reaching
$\sim\negmedspace10\%\ \text{MR}$ our FasterRCNN is the simplest.

\paragraph{Conclusion}

Once properly adapted, FasterRCNN obtains competitive performance
for pedestrian detection on the Caltech dataset. This is the model
we will use in all following experiments.

In section \ref{sec:CityPersons} we introduce a new dataset that
will enable further improvements of detection performance.

\section{\label{sec:CityPersons}CityPersons dataset}

The Cityscapes dataset \cite{Cordts2016Cityscapes} was created for
the task of semantic segmentation in urban street scenes. It consists
of a large and diverse set of stereo video sequences recorded in streets
from different cities in Germany and neighbouring countries. Fine
pixel-level annotations of 30 visual classes are provided for $5\,000$
images from 27 cities. The fine annotations include instance labels
for persons and vehicles. Additionally $20\,000$ images from 23 other
cities are annotated with coarse semantic labels, without instance
labels.

In this paper, we present the \texttt{\small{}CityPersons} dataset,
built upon the Cityscapes data to provide a new dataset of interest
for the pedestrian detection community. For each frame in the $5\,000$
fine-annotations subset, we have created high quality bounding box
annotations for pedestrians (section \ref{subsec:bb-annos}). In section
\ref{subsec:annotations-statistics} we contrast \texttt{\small{}CityPersons}
with previous datasets regarding: volume, diversity and occlusion.
In section \ref{sec:Improve-quality} we show how to use this new
data to improve results on other datasets.

\subsection{\label{subsec:bb-annos}Bounding box annotations}

The Cityscapes dataset already provides instance level segments for
each human. These segments indicate the visible parts of humans. Simply
using bounding boxes of these segments would raise three issues. I1)
The box aspect ratio would be irregular, persons walking have varying
width. It has been proposed to thus normalize aspect ratio for pedestrian
annotations. I2) Even after normalizing aspect ratio, the boxes would
not align amongst each other. They will be off in the horizontal axis
due to being normalized based on the segment centre rather the object
centre. They will be off in the vertical axis due to variable level
of occlusion for each person. It has been shown that pedestrian detectors
benefit from well aligned training samples \cite{shanshan_cvpr16},
and conversely, training with misaligned samples will hamper results.
I3) Existing datasets (INRIA, Caltech, KITTI) have defined bounding
boxes covering the full object extent, not just the visible area.
In order to train compatible, high quality models, we need to have
annotations that align well the full extent of the persons bodies
(``amodal bounding box'' \cite{Li2016Eccv}).

\paragraph{Fine-grained categories}

In the Cityscapes dataset, humans are labelled as either person or
rider. In this paper, we provide further fine-grained labels for persons.
Based on the postures, we group all humans into four categories: pedestrian
(walking, running or standing up), rider (riding bicycles or motorbikes),
sitting person, and other person (with unusual postures, e.g. stretching).

\paragraph{Annotation protocol}

\begin{figure}
\begin{centering}
\subfloat[Image]{\begin{centering}
\includegraphics[width=0.3\columnwidth]{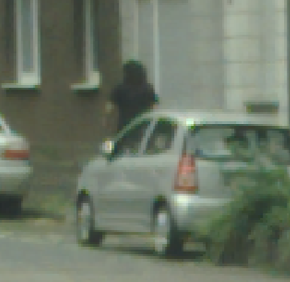}
\par\end{centering}
}~\subfloat[Segmentation mask]{\begin{centering}
\includegraphics[width=0.3\columnwidth]{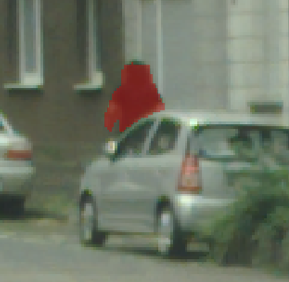}
\par\end{centering}
}~\subfloat[Bounding box anno.]{\begin{centering}
\includegraphics[width=0.3\columnwidth]{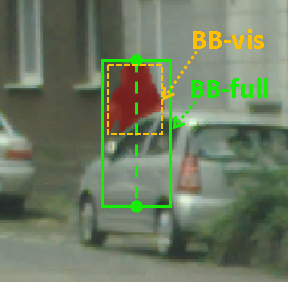}
\par\end{centering}
}
\par\end{centering}
\centering{}\caption{\label{fig:Examples-bb-annos}Illustration of bounding box annotations
for pedestrians. For each person, the top of the head and middle of
the feet is drawn by the annotator. An aligned bounding box is automatically
generated using the fixed aspect ratio (0.41). The bounding box covering
the segmentation mask is used to estimate the visible part.}
\end{figure}

For pedestrians and riders (cyclists, motorists), we follow the same
protocol as used in \cite{shanshan_cvpr16}, where the full body is
annotated by drawing a line from the top of the head to the middle
of two feet, and the bounding box is generated using a fixed aspect
ratio (0.41). This protocol has been shown to provide accurate alignments.
The visible bounding box for each instance is the tightest one fully
covering the segment mask, and can be generated automatically from
the segment. See an illustration in figure \ref{fig:Examples-bb-annos}.
The occlusion ratio can then be computed as $\frac{area(BB-vis)}{area(BB-full)}$.

As of other categories of persons, i.e. sitting and other persons,
there is no uniform alignment to apply, so we only provide the segment
bounding box for each of them without full body annotations. 

Apart from real persons, we also ask the annotators to search over
the whole image for areas containing fake humans, for instance, people
on posters, statue, mannequin, people's reflection in mirror or window,
etc., and mark them as ignore regions.

\paragraph{Annotation tool}

Since we already have the segment mask for each instance, we can do
the annotations in a more efficient way than from scratch. To this
end, we develop a new annotation tool to avoid searching for persons
over the images by exploiting the available instance segments. This
tool pops out one person segment at a time and asks the annotator
to recognize the fine-grained category first and then do the full
body annotation for pedestrians and riders. Thanks to the high-quality
of segmentation annotations, using such a tool also reduces the risk
of missing persons, especially at crowded scenes. But the ignore region
annotations have to be done by searching over the whole images.

\subsection{\label{subsec:annotations-statistics}Statistics}

\paragraph{Volume}

We show the number of bounding box annotations provided by us in table
\ref{tab:Statistics-bb}. In a total of $5\,000$ images, we have
\textasciitilde{}35k person and \textasciitilde{}13k ignore region
annotations. And we notice the density of persons are consistent across
train/validation/test subsets. Please note we use the same split as
Cityscapes. 

\begin{table}
\begin{centering}
\begin{tabular}{l|ccc|c}
 & Train & Val. & Test & Sum\tabularnewline
\hline 
\hline 
\multirow{1}{*}{\# cities} & 18 & 3 & 6 & 27\tabularnewline
\multirow{1}{*}{\# images} & $2\,975$ & $500$ & $1\,575$ & $5\,000$\tabularnewline
\hline 
\# persons & $19\,654$ & $3\,938$ & $11\,424$ & $35\,016$\tabularnewline
\multirow{1}{*}{\# ignore regions} & $6\,768$ & $1\,631$ & $4\,773$ & $13\,172$\tabularnewline
\end{tabular}
\par\end{centering}
\caption{\label{tab:Statistics-bb}Statistics of bounding box annotations on
\texttt{\small{}CityPersons} dataset.}

\end{table}

\paragraph{Diversity}

We compare the diversity of Caltech, KITTI and \texttt{\small{}CityPersons}
in table \ref{tab:statistics-comparison}. Since KITTI test set annotations
are not publicly available, we only consider the training subset for
a fair comparison.

The \texttt{\small{}CityPersons} training subset was recorded across
18 different cities, three different seasons, and various weather
conditions. While the Caltech and KITTI datasets are only recorded
in one city at one season each. 

In terms of density, we have on average \textasciitilde{}7 persons
per image. This number is much higher than that on the Caltech and
KITTI datasets, where each image only contains \textasciitilde{}1
person on average.

Also, the number of identical persons is another important evidence
of diversity. On our \texttt{\small{}CityPersons} dataset, the number
of identical persons amounts up to $\sim\negmedspace20\,000$. In
contrast, the Caltech and KITTI dataset only contains \textasciitilde{}
$1\,300$ and \textasciitilde{} $6\,000$ unique pedestrians respectively.
Note KITTI and \texttt{\small{}CityPersons} frames are sampled very
sparsely, so each person is considered as unique.

\begin{figure}
\begin{centering}
\includegraphics[width=0.3\textwidth]{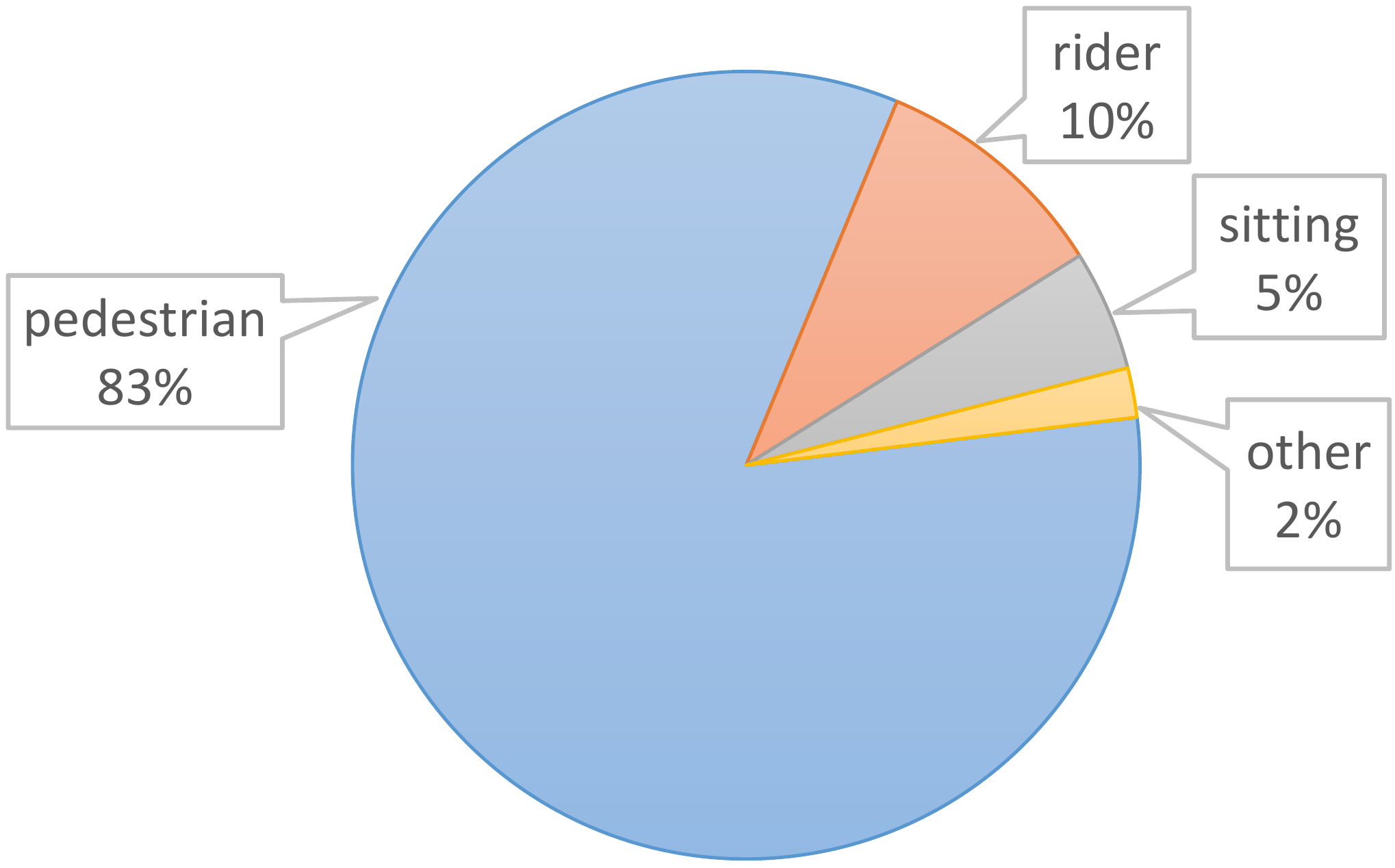}
\par\end{centering}
\caption{\label{fig:human-categories}Fine-grained person categories on \texttt{\small{}CityPersons}.}
\end{figure}

\texttt{\small{}CityPersons} also provides fine-grained labels for
persons. As shown in figure \ref{fig:human-categories}, pedestrians
are the majority (83\%). Although riders and sitting persons only
occupy 10\% and 5\% respectively, the absolute numbers are still considerable,
as we have a large pool of \textasciitilde{}35k persons.

\begin{table}
\begin{centering}
\begin{tabular}{l|c|c|c}
 & Caltech & KITTI & CityPersons\tabularnewline
\hline 
\hline 
\# country & 1 & 1 & 3\tabularnewline
\# city & 1 & 1 & 18\tabularnewline
\# season & 1 & 1 & 3\tabularnewline
\# person/image & 1.4 & 0.8 & 7.0\tabularnewline
\# unique person & $1\,273$ & $6\,336$ & $19\,654$\tabularnewline
\end{tabular}
\par\end{centering}
\caption{\label{tab:statistics-comparison}Comparison of diversity on different
datasets (training subset only).}
\end{table}

\paragraph{Occlusion}

\begin{figure}
\begin{centering}
\includegraphics[width=0.5\textwidth]{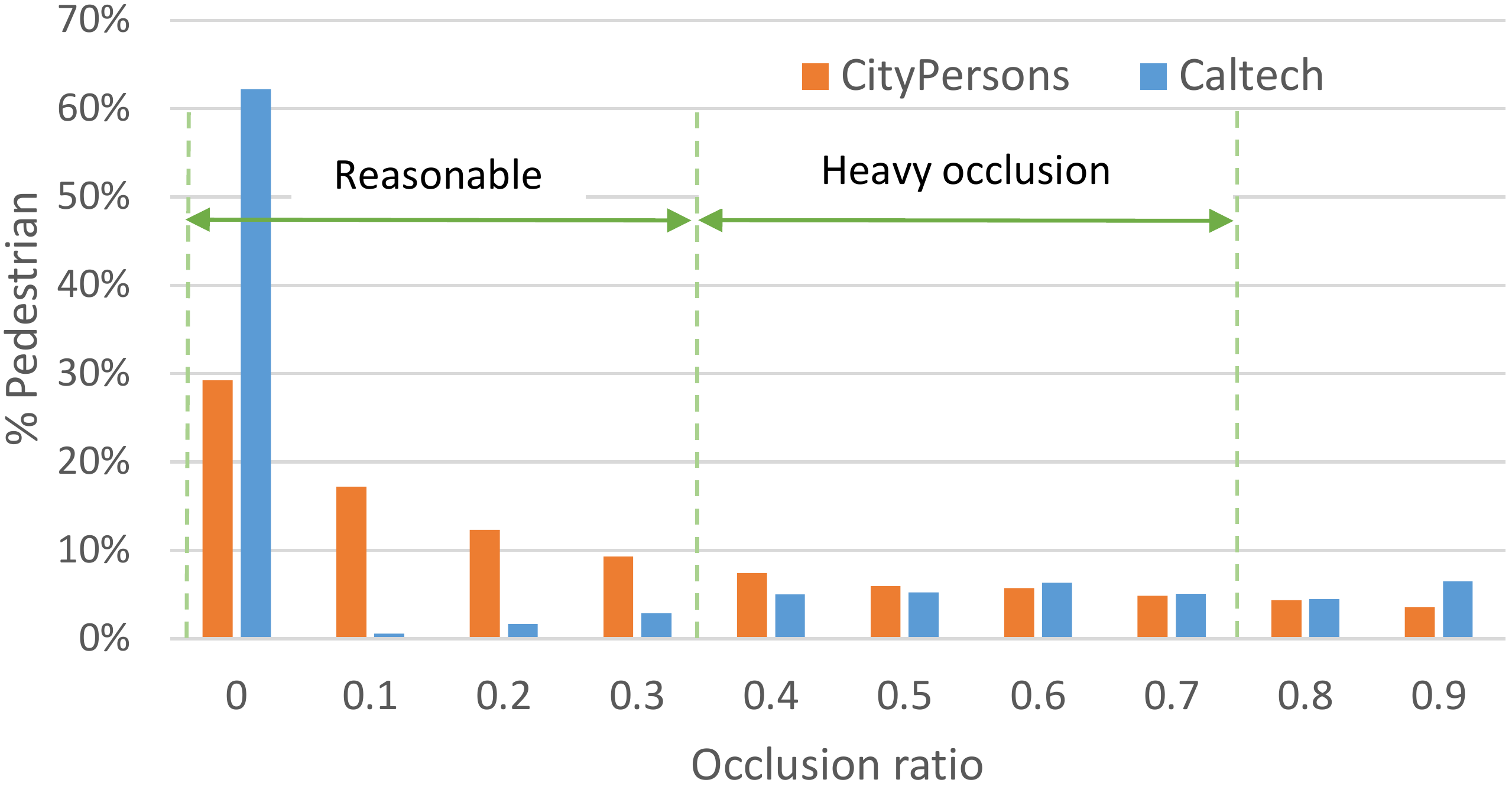}
\par\end{centering}
\caption{\label{fig:Occlusion-distribution}Comparison of occlusion distributions
on \texttt{\small{}CityPersons} and Caltech datasets. \texttt{\small{}CityPersons}
contains more occlusions in the reasonable subset than Caltech.}
\end{figure}

\begin{figure}
\begin{centering}
\includegraphics[width=1\linewidth]{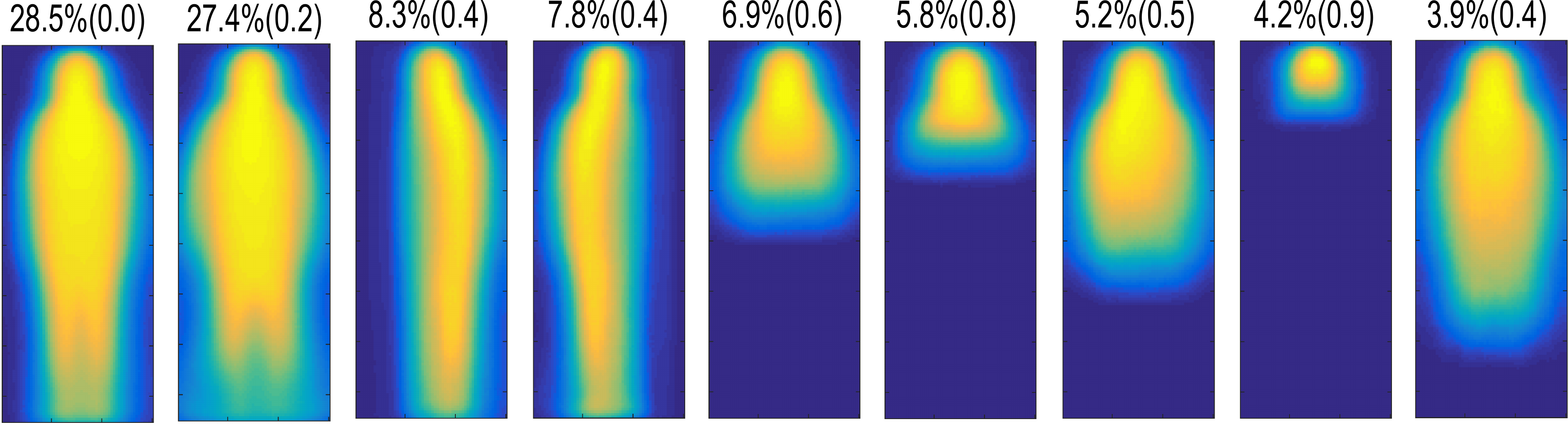}
\par\end{centering}
\caption{\label{fig:Occlusion-patterns}Top 9 of quantized 11 occlusion patterns
of pedestrians on \texttt{\small{}CityPersons} dataset. Two numbers
on top indicate percentage and average occlusion ratio of samples
clustered into each pattern.}
\end{figure}

The Cityscapes data was collected by driving through the centre of
some highly populated cities, e.g. Frankfurt and Hamburg. We notice
that on some images, there are \textasciitilde{}100 people walking
on the street, highly occluded by each other. Such a high occlusion
is rarely seen in previous datasets. In figure \ref{fig:Occlusion-distribution},
we compare the distribution of pedestrians at different occlusion
levels for Caltech and \texttt{\small{}CityPersons}. We notice that
on Caltech there are more than 60\% fully visible pedestrians, while
on \texttt{\small{}CityPersons} there are less than 30\%. This indicates
we have two times more occlusions than Caltech, which makes \texttt{\small{}CityPersons}
a more interesting ground for occlusion handling. Moreover, on the
reasonable subset (<=0.35 occlusion) the community typically use,
Caltech is dominated by fully visible pedestrians, while \texttt{\small{}CityPersons}
has more occlusion cases.

In order to understand which kinds of occlusions we have on \texttt{\small{}CityPersons},
we quantize all persons into 11 patterns and show the top 9 of them
in figure \ref{fig:Occlusion-patterns} (the last two patterns are
not shown as they are of less than 1\% and thus noisy). For visualization,
we resize each full body bounding box to a fixed size, and then overlay
the segmentation mask. For each pattern, the bright area shows the
visible part and the two numbers on top indicate the percentage and
average occlusion ratio of corresponding pattern. The first two patterns
(55.9\%) roughly cover the ``reasonable'' subset; the third and
fourth patterns correspond to occlusions from either left or right
side. Apart from that, we still have about 30\% pedestrians distributed
in various patterns, some of which have a very high occlusion ratio
(>0.9). Such distributed occlusion patterns increase the diversity
of the data and hence makes the dataset a more challenging test base.

\subsection{Benchmarking}

With the publication of this paper, we will create a website for \texttt{\small{}CityPersons}
dataset, where train/validation annotations can be downloaded, and
an online evaluation server is available to compute numbers over the
held-out test annotations.\footnote{As a subset of the Cityscapes dataset,\texttt{\small{} CityPersons}
annotations and benchmark will be available on the Cityscapes website.
The evaluation server is being setup and the metrics will change.}

We follow the same evaluation protocol as used for Caltech \cite{Dollar2012Pami},
by allowing evaluation on different subsets. In this paper, $\mbox{MR}$
stands for log-average miss rate on the ``reasonable'' setup (scale
{[}{\small{}$50,\,\infty$}{]}, occlusion ratio {[}0, 0.35{]}) unless
otherwise specified. While evaluating pedestrian detection performance,
cyclists/sitting persons/other persons/ignore regions are not considered,
which means detections matching with those areas are not counted as
mistakes.

\subsection{\label{sec:Baseline-experiments}Baseline experiments}

To understand the difficulties of pedestrian detection on the \texttt{\small{}CityPersons}
dataset, we train and evaluate three different detectors. ACF \cite{Dollar2014Pami}
and Checkerboards \cite{Zhang2015Cvpr} are representatives from the
Integral Channel Features detector (ICF) family, while FasterRCNN
\cite{renNIPS15fasterrcnn} acts as the state-of-the-art detector.
We set up the FasterRCNN detector by following the practices we learned
from Caltech experiments (section \ref{subsec:faster-rcnn}). Since
\texttt{\small{}CityPersons} images are \textasciitilde{}7 times larger
than Caltech, we are only able to use an upsampling factor of 1.3
to fit in 12GB of GPU memory.

We re-train each detector using the \texttt{\small{}CityPersons} training
set and then evaluate on the validation set. Note that all the \texttt{\small{}CityPersons}
numbers reported in this paper are on the validation set. Consistent
with the reasonable evaluation protocol, we only use the reasonable
subset of pedestrians for training; cyclists/sitting persons/other
persons/ignore regions are avoided for negative sampling.

\begin{figure}
\begin{centering}
\includegraphics[width=0.45\textwidth]{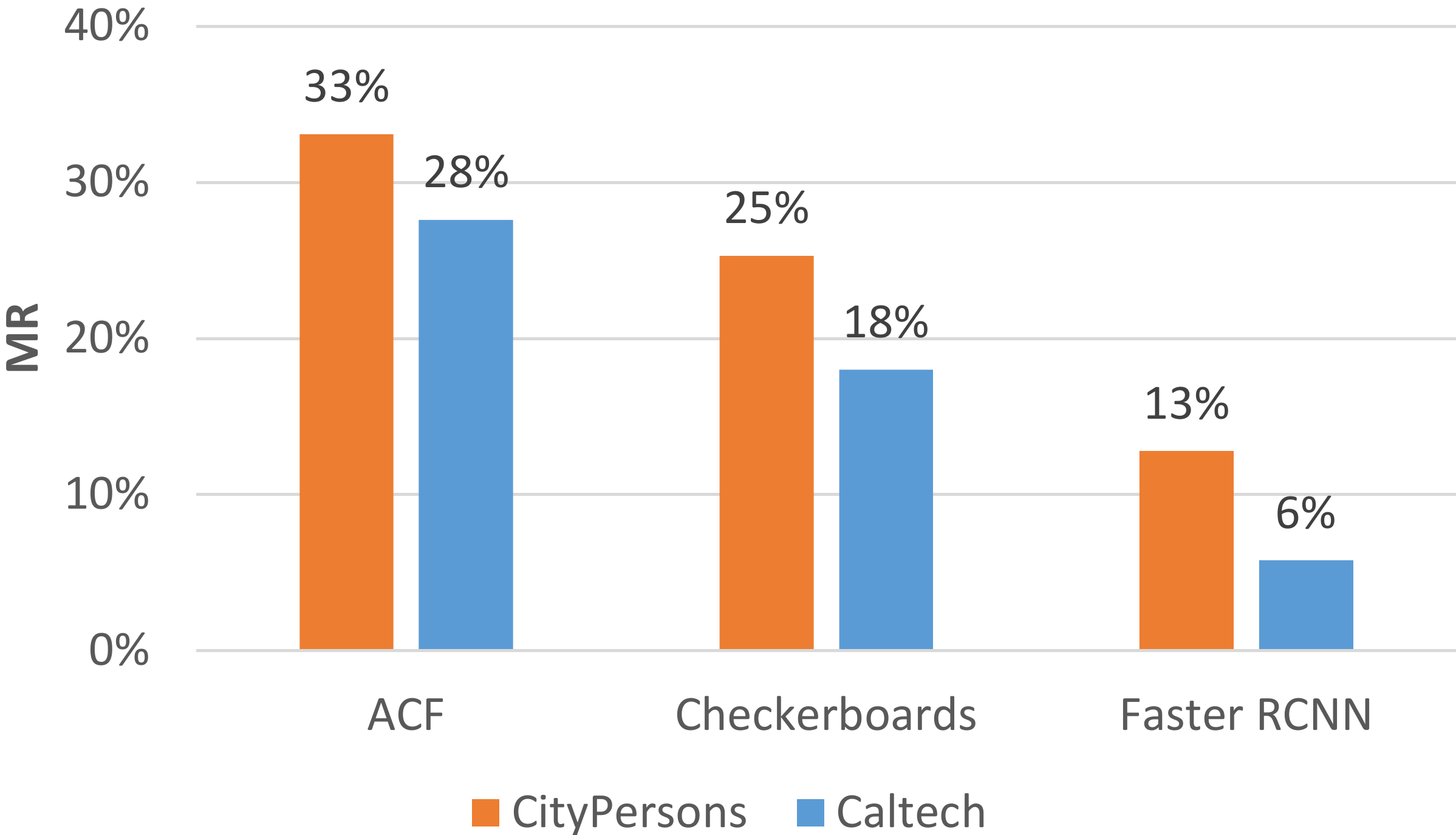}
\par\end{centering}
\caption{\label{fig:Baseline-experiments}Comparison of baseline detectors
on Caltech test and \texttt{\small{}CityPersons} val. set (reasonable).
Numbers are $\mbox{MR}^{N}$ on Caltech and MR on \texttt{\small{}CityPersons}
(lower is better). Ranking of methods between two datasets is stable.
For all methods, \texttt{\small{}CityPersons} is more difficult to
solve than Caltech.}

\end{figure}

In figure \ref{fig:Baseline-experiments}, we show the comparison
of the above three detectors on \texttt{\small{}CityPersons} and Caltech.
FasterRCNN outperforms ICF detectors by a large margin, which indicates
the adaptation of FasterRCNN on Caltech is also transferable to \texttt{\small{}CityPersons}.
Moreover, we find the ranking of three detectors on \texttt{\small{}CityPersons}
is consistent with that on Caltech, but the performance on \texttt{\small{}CityPersons}
dataset is lower for all three detectors. This comparison shows that
\texttt{\small{}CityPersons} is a more challenging dataset, thus more
interesting for future research in this area. 

\begin{figure}
\begin{centering}
\includegraphics[width=0.4\textwidth]{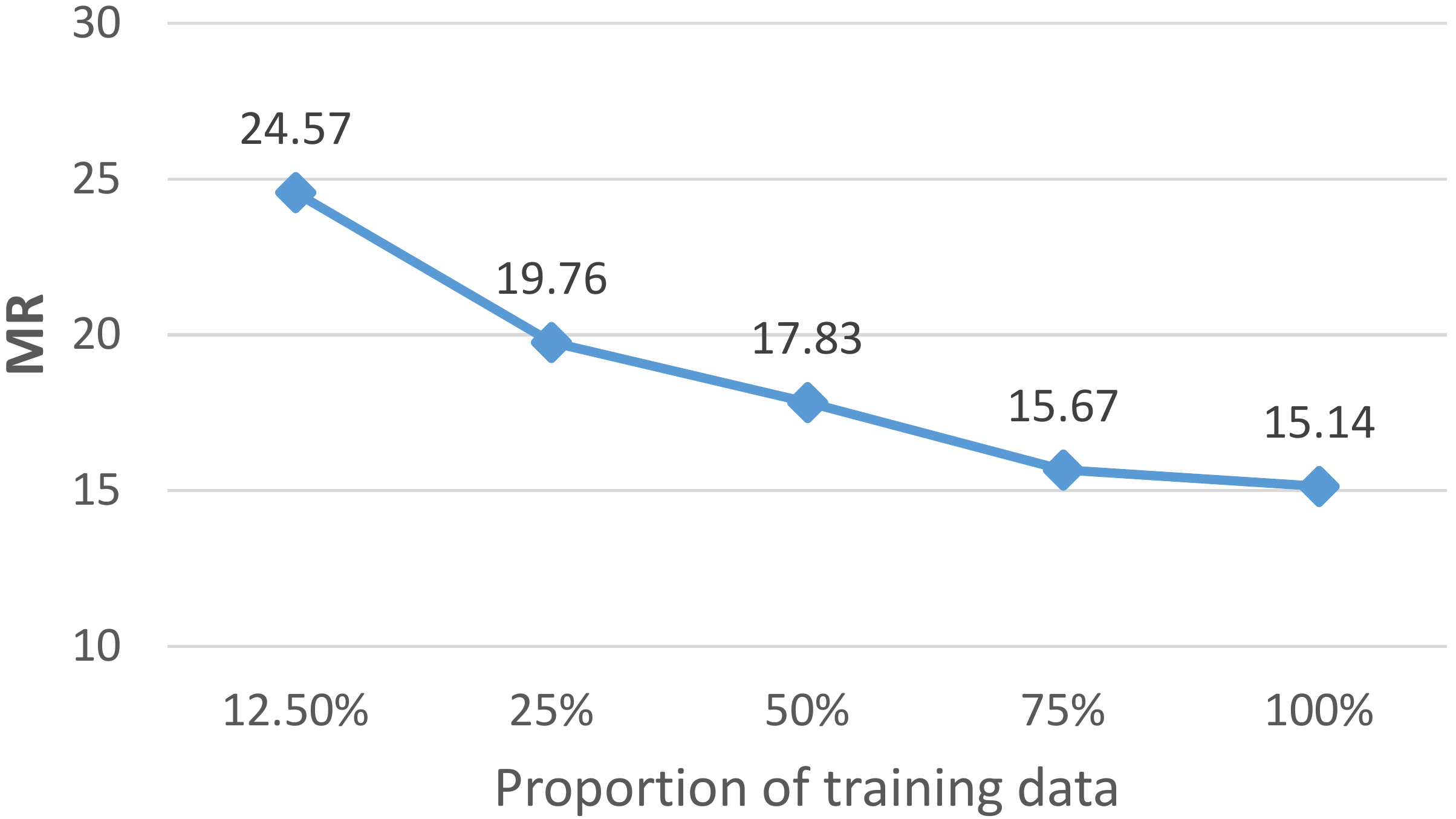}
\par\end{centering}
\caption{\label{fig:proportion-training-data}Quality as function of training
volume. Fast\-er\-RCNN model trained/evaluated on \texttt{\small{}CityPersons}
train/val. set (MR: lower is better).}
\end{figure}

To understand the impact of having a larger amount of training data,
we show how performance grows as training data increases in figure
\ref{fig:proportion-training-data}. We can see performance keeps
improving with more data. Therefore, it is of great importance to
provide CNNs with a large amount of data.

Considering the trade off between speed and quality, we use an alternative
model of our FasterRCNN by switching off input image upsampling for
the analysis experiments shown in figure \ref{fig:proportion-training-data}
and section \ref{subsec:Semantic-labels}. This model is about 2x
faster at both training and test time, but only drops the performance
by \textasciitilde{}2 pp (from 13\% MR to 15\% MR).

\paragraph{Conclusion}

The \texttt{\small{}CityPersons} dataset can serve as a large and
diverse database for training a powerful model, as well as a more
challenging test base for future research on pedestrian detection. 

\section{\label{sec:Improve-quality}Improve quality using CityPersons}

Having the \texttt{\small{}CityPersons} dataset at hand, we now proceed
to illustrate three different ways it enables to improve pedestrian
detection results (\S\ref{subsec:Generalisation-across-datasets},
\S\ref{subsec:Better-pre-training-improves-quality}, \S\ref{subsec:Semantic-labels}).
As we will see, \texttt{\small{}CityPersons} is particularly effective
at improving results for small scale pedestrians, occluded ones, and
providing higher localization accuracy.

\subsection{\label{subsec:Generalisation-across-datasets}Generalization across
datasets}

Commonly, a detector is trained on the training set of the target
benchmark. As such, one needs to train multiple detectors for different
benchmarks. Ideally, one would wish to train one detector that is
able to perform well on multiple benchmarks. Since the \texttt{\small{}CityPersons}
dataset is large and diverse, we wonder if it can allow us to train
a detector with good generalization capabilities.

To see how well \texttt{\small{}CityPersons} data generalizes across
different datasets, we train models on Caltech, KITTI and \texttt{\small{}CityPersons}
datasets, and then apply each of them on six different test sets:
Caltech, KITTI, \texttt{\small{}CityPersons}, INRIA, ETH and Tud-Brussels.
For KITTI, we split the public training data into training and validation
subsets (2:1) by random sampling. Table \ref{tab:Generalisation-ability}
shows comparisons of two detectors: ACF \cite{Dollar2014Pami} and
FasterRCNN \cite{renNIPS15fasterrcnn}. 

We observe:

(1) Overall, when trained with the same data FasterRCNN generalizes
better across datasets than ACF. (Note that FasterRCNN benefits from
ImageNet pre-training, while ACF does not.)

(2) For both detectors, the mean MR across test sets is significantly
better for models trained with \texttt{\small{}CityPersons} training
data. \texttt{\small{}CityPersons} generalizes better than Caltech
and KITTI.

\begin{table}
\begin{centering}
\subfloat[ACF]{\centering{}%
\begin{tabular}{cc|c|c|c}
 & Train & \multirow{2}{*}{Caltech} & \multirow{2}{*}{KITTI} & \multirow{2}{*}{CityPersons}\tabularnewline
Test &  &  &  & \tabularnewline
\hline 
\hline 
\multicolumn{2}{c|}{Caltech} & \textit{27.63} & 63.15 & \textbf{51.28}\tabularnewline
\multicolumn{2}{c|}{KITTI} & 49.99 & \textit{32.06} & \textbf{46.74}\tabularnewline
\multicolumn{2}{c|}{CityPersons} & \textbf{72.89} & 94.28 & \textit{33.10}\tabularnewline
\hline 
\multicolumn{2}{c|}{INRIA} & 63.39 & 67.49 & \textbf{50.23}\tabularnewline
\multicolumn{2}{c|}{ETH} & 78.64 & 89.94 & \textbf{56.30}\tabularnewline
\multicolumn{2}{c|}{Tud-Brussels} & \textbf{63.22} & 69.25 & 67.21\tabularnewline
\hline 
\multicolumn{2}{c|}{mean MR} & 59.29 & 69.36 & \textbf{50.81}\tabularnewline
\end{tabular}}
\par\end{centering}
\begin{centering}
\subfloat[FasterRCNN]{\centering{}%
\begin{tabular}{cc|c|c|c}
 & Train & \multirow{2}{*}{Caltech} & \multirow{2}{*}{KITTI} & \multirow{2}{*}{CityPersons}\tabularnewline
Test &  &  &  & \tabularnewline
\hline 
\hline 
\multicolumn{2}{c|}{Caltech} & \textit{10.27} & 46.86 & \textbf{21.18}\tabularnewline
\multicolumn{2}{c|}{KITTI} & 10.50 & \textit{8.37} & \textbf{8.67}\tabularnewline
\multicolumn{2}{c|}{CityPersons} & \textbf{46.91} & 51.21 & \textit{12.81}\tabularnewline
\hline 
\multicolumn{2}{c|}{INRIA} & 11.47 & 27.53 & \textbf{10.44}\tabularnewline
\multicolumn{2}{c|}{ETH} & 57.85 & 49.00 & \textbf{35.64}\tabularnewline
\multicolumn{2}{c|}{Tud-Brussels} & 42.89 & 45.28 & \textbf{36.98}\tabularnewline
\hline 
\multicolumn{2}{c|}{mean MR} & 29.98 & 38.04 & \textbf{20.95}\tabularnewline
\end{tabular}}
\par\end{centering}
\caption{\label{tab:Generalisation-ability}Generalization ability of two different
methods, trained and tested over different datasets. All numbers are
$\text{MR}$ on reasonable subset. Bold indicates the best results
obtained via generalization across datasets (different train and test).}
\end{table}

These experiments confirm the generalization ability of \texttt{\small{}CityPersons}
dataset, that we attribute to the size and diversity of the Cityscapes
data, and to the quality of the bounding boxes annotations.

\subsection{\label{subsec:Better-pre-training-improves-quality}Better pre-training
improves quality}

In table \ref{tab:Generalisation-ability}, we find the \texttt{\small{}CityPersons}
data acts as very good source of training data for different datasets,
assuming we are blind to the target domain. Furthermore, when we have
some training data from the target domain, we show \texttt{\small{}CityPersons}
data can be also used as effective external training data, which helps
to further boost performance.

First, we consider Caltech as the target domain, and compare the quality
of two models. One is trained on Caltech data only; and the other
is first trained on \texttt{\small{}CityPersons}, and then finetuned
on Caltech (\texttt{\small{}CityPersons}$\rightarrow$Caltech). From
table \ref{tab:gain-from-citypersons-caltech}, we can see the additional
training with \texttt{\small{}CityPersons} data improves the performance
in the following three aspects.

(1) \texttt{\small{}CityPersons} data improves overall performance.
When evaluated on the reasonable setup, the \texttt{\small{}CityPersons}$\rightarrow$Caltech
model obtains \textasciitilde{}1 pp gain.

(2) \texttt{\small{}CityPersons} data improves more for harder cases,
e.g. smaller scale, heavy occlusion. We notice the gap for heavy occlusion
is large (\textasciitilde{}9 pp), due to more occluded training samples
on the \texttt{\small{}CityPersons} dataset. Similar trend is also
found for smaller scale persons ({[}30,80{]}).

(3) \texttt{\small{}CityPersons} data helps to produce better-aligned
detections. The Caltech new annotations are well aligned, thus a good
test base for alignment quality of detections. When we increase the
IoU threshold for matching from 0.50 to 0.75, the gain from \texttt{\small{}CityPersons}
data also grows from 1 pp to 5 pp. This gap indicates the high quality
of \texttt{\small{}CityPersons} annotations are beneficial to produce
better-aligned detections. 

\begin{figure}
\begin{centering}
\subfloat[\label{fig:IoU=00003D0.50}IoU=0.50]{\centering{}\includegraphics[width=0.4\textwidth]{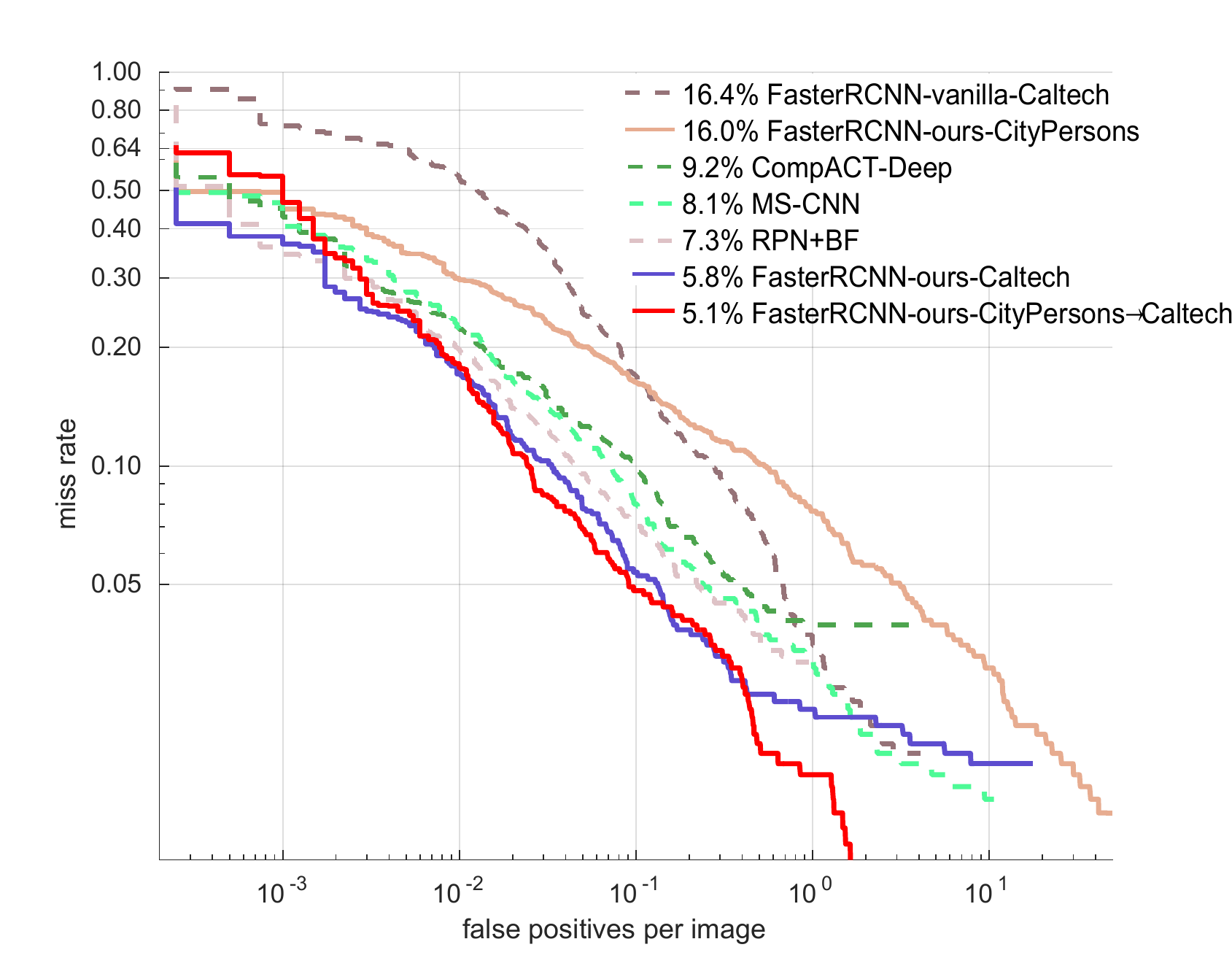}}
\par\end{centering}
\begin{centering}
\subfloat[\label{fig:IoU=00003D0.75}IoU=0.75]{\begin{centering}
\includegraphics[width=0.4\textwidth]{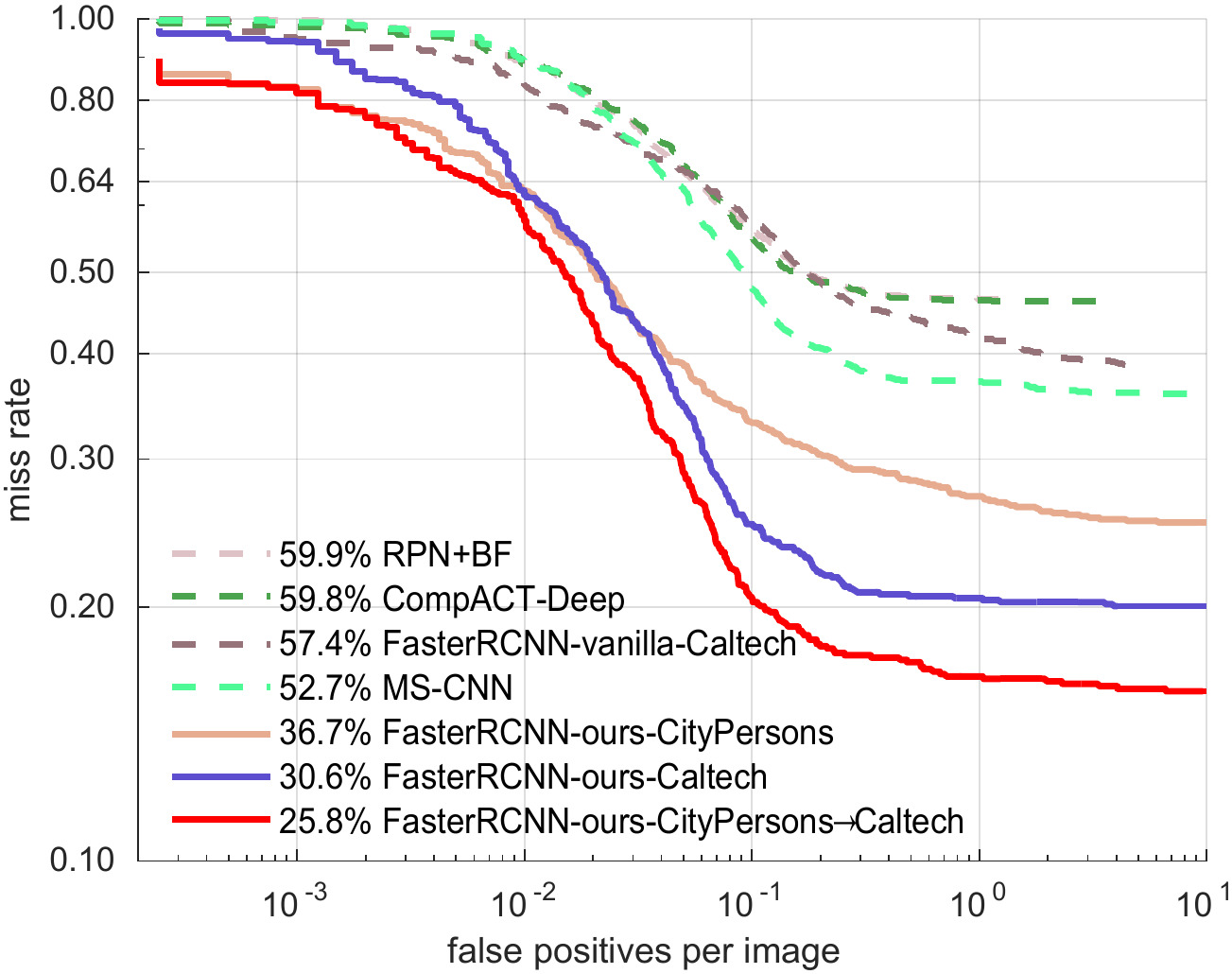}
\par\end{centering}
}
\par\end{centering}
\caption{\label{fig:caltech-curves}Comparison of state-of-the-art results
on the Caltech test set (reasonable subset), $\text{MR}^{N}$.}
\end{figure}

Compared with other state-of-the-art detectors, our best model using
\texttt{\small{}CityPersons} for pre-training obtains 5.1\% $\mbox{MR}^{N}$
at IoU 0.50 evaluation, outperforming previous best reported results
(7.3\% $\mbox{MR}^{N}$) by 2.2 pp (figure \ref{fig:IoU=00003D0.50});
this gap becomes even larger (\textasciitilde{} 20 pp) when we use
a stricter IoU of 0.75 (figure \ref{fig:IoU=00003D0.75}). From the
comparison, our FasterRCNN detector obtains state-of-the-art results
on Caltech, and improves the localization quality significantly.

\begin{table}
\centering{}\hspace*{-1em}\setlength{\tabcolsep}{4pt} 
\begin{tabular}{clcc|cc|c}
\multirow{2}{*}{$\nicefrac{O}{N}$} & \multirow{2}{*}{Setup} & {\small{}Scale} & \multirow{2}{*}{IoU} & Cal- & {\small{}CityPersons } & \multirow{2}{*}{$\Delta\mbox{MR}$}\tabularnewline
 &  & {\small{}range} &  & tech & {\small{}$\rightarrow$Caltech} & \tabularnewline
\hline 
$\mbox{MR}^{O}$ & {\small{}Reasonable} & {\small{}$\left[50,\,\infty\right]$} & 0.5 & 10.3 & 9.2 & + 1.1\tabularnewline
$\mbox{MR}^{O}$ & {\small{}Smaller} & {\small{}$\left[30,\,80\right]$} & 0.5 & 52.0 & 48.5 & \textbf{+ 3.5}\tabularnewline
$\mbox{MR}^{O}$ & {\small{}Heavy occl.} & {\small{}$\left[50,\,\infty\right]$} & 0.5 & 68.3 & 57.7 & \textbf{+ 8.6}\tabularnewline
\hline 
$\mbox{MR}^{N}$ & {\small{}Reasonable} & {\small{}$\left[50,\,\infty\right]$} & 0.5 & 5.8 & 5.1 & + 0.7\tabularnewline
$\mbox{MR}^{N}$ & {\small{}Reasonable} & {\small{}$\left[50,\,\infty\right]$} & 0.75 & 30.6 & 25.8 & \textbf{+ 4.8}\tabularnewline
\end{tabular}\caption{\label{tab:gain-from-citypersons-caltech}Gains from additional \texttt{\small{}CityPersons}
training at different evaluation setups on Caltech test set. $\mbox{MR}^{O}$
and $\mbox{MR}^{N}$ indicate numbers evaluated on original and new
annotations \cite{shanshan_cvpr16}. \texttt{\small{}CityPersons}
pre-training helps more for more difficult cases. See also table \ref{tab:gain-from-persons-kitti}.}
\end{table}

\begin{figure}
\begin{centering}
\subfloat[Original image]{\centering{}\includegraphics[width=0.23\textwidth]{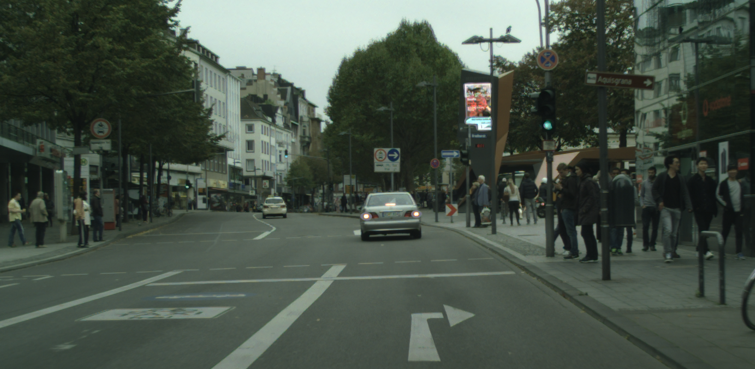}}
\ \subfloat[Semantic map]{\centering{}\includegraphics[width=0.23\textwidth]{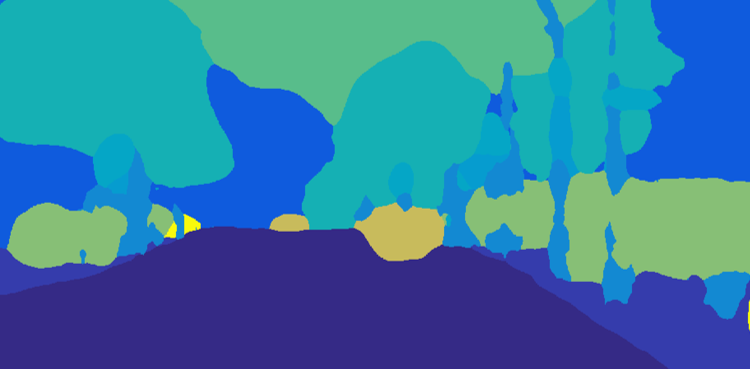}}
\par\end{centering}
\caption{\label{fig:Example-semantic-labels}Example of semantic map generated
by an FCN-8s model trained on Cityscapes coarse annotations.}
\end{figure}

When we consider KITTI as the target domain, we also see improvements
brought by additional training with \texttt{\small{}CityPersons} data.
As shown in table \ref{tab:gain-from-persons-kitti}, the gain on
reasonable evaluation setup is 2.5 pp, while for smaller scale, the
gap becomes more impressive (10.7 pp). The 4.1 pp gap at IoU 0.75
again verifies \texttt{\small{}CityPersons} data helps to produce
better aligned detections.

\begin{table}
\centering{}\setlength{\tabcolsep}{4pt} 
\begin{tabular}{lcc|cc|c}
\multirow{2}{*}{Setup} & {\small{}Scale} & \multirow{2}{*}{IoU} & \multirow{2}{*}{{\small{}KITTI}} & {\small{}CityPersons} & \multirow{2}{*}{$\Delta\mbox{MR}$}\tabularnewline
 & {\small{}range} &  &  & {\small{}$\rightarrow$KITTI} & \tabularnewline
\hline 
\hline 
Reasonable & {\small{}$\left[50,\,\infty\right]$} & 0.5 & 8.4 & 5.9 & + 2.5\tabularnewline
Reasonable & {\small{}$\left[50,\,\infty\right]$} & 0.75 & 43.3 & 39.2 & \textbf{+ 4.1}\tabularnewline
Smaller & {\small{}$\left[30,\,80\right]$} & 0.5 & 37.8 & 27.1 & \textbf{+ 10.7}\tabularnewline
\end{tabular}\caption{\label{tab:gain-from-persons-kitti}Gains from additional \texttt{\small{}CityPersons}
training at different evaluation setups on KITTI validation set. All
numbers are $\text{MR}$ (see \S\ref{sec:pedestrian-convnet}). Here
also, \texttt{\small{}CityPersons} pre-training helps more for more
difficult cases. See also table \ref{tab:gain-from-citypersons-caltech}.}
\end{table}

\subsection{\label{subsec:Semantic-labels}Exploiting Cityscapes semantic labels}

In this subsection, we explore how much improvement can be obtained
for pedestrian detection by leveraging the semantic labels available
on the Cityscapes dataset. 

We use an FCN-8s \cite{FCN15cvpr} model trained on Cityscapes coarse
annotations to predict semantic labels. Note we cannot involve fine-annotation
images in this semantic labelling training, otherwise our following
detection training will suffer from overfitting. Although this model
is only trained on coarse annotations, we can see the semantic segmentation
mask provides a reasonable structure for the whole scene (figure \ref{fig:Example-semantic-labels}).
Then we concatenate semantic channels with RGB channels and feed them
altogether into convnets, letting convnets to figure out the hidden
complementarity. 

For the reasonable evaluation setup, we get an overall improvement
of \textasciitilde{}0.6 pp from semantic channels. When we look at
the fine-grained improvements for different scale ranges, we find
that semantic channels help more for small persons, which is a hard
case for our task (table \ref{tab:Improvements-semantic}). 

As a preliminary trial, we already get some improvements from semantic
labels, which encourage us to explore more effective ways of using
semantic information.

\begin{table}
\centering{}%
\begin{tabular}{c|c|c|c}
Scale range & Baseline & + Semantic & $\Delta\mbox{MR}$\tabularnewline
\hline 
\hline 
$\left[50,\hspace{0.7em}\infty\right]$ & 15.4 & 14.8 & + 0.6\tabularnewline
\hline 
$\left[100,\,\infty\right]$ & 7.9 & 8.0 & + 0.1\tabularnewline
$\left[75,\,100\right]$ & 7.2 & 6.7 & + 0.5\tabularnewline
$\left[50,\hspace{0.7em}75\right]$  & 25.6 & 22.6 & \textbf{+ 3.0}\tabularnewline
\end{tabular}\caption{\label{tab:Improvements-semantic}Improvements from semantic channels
in different scale ranges. Numbers are MR on the \texttt{\small{}CityPersons}
val. set. Albeit there is small overall gain, adding semantic channels
helps for the difficult case of small persons.}
\end{table}

\section{\label{sec:Conclusion}Summary}

In this paper, we first show that a properly adapted FasterRCNN can
achieve state-of-the-art performance on Caltech. Aiming for further
improvement from more and better data, we propose a new diverse dataset
namely \texttt{\small{}CityPersons} by providing bounding box annotations
for persons on top of Cityscapes dataset. \texttt{\small{}CityPersons}
shows high contrast to previous datasets as it consists of images
recorded across 27 cities, 3 seasons, various weather conditions and
more common crowds. 

Serving as training data, \texttt{\small{}CityPersons} shows strong
generalization ability from across dataset experiments. Our FasterRCNN
model trained on \texttt{\small{}CityPersons} obtains reasonable performance
over six different benchmarks. Moreover, it further improves the detection
performance with additional finetuning on the target data, especially
for harder cases (small scale and heavy occlusion), and also enhance
the localization quality.

On the other hand, \texttt{\small{}CityPersons} can also be used as
a new test benchmark as there are more challenges, e.g. more occlusions
and diverse environments. We will create a website for this benchmark
and only allows for online evaluations by holding out the test set
annotations.

Other than bounding box annotations for persons, there are additional
information to leverage on \texttt{\small{}CityPersons}, for instance,
fine semantic segmentations, other modalities of data (stereo, GPS),
and un-annotated neighbouring frames. Our preliminary results of using
semantic labels show promising complementarity. These rich data will
motivate more efforts to solve the problem of pedestrian detection.

\bibliographystyle{ieee}
\bibliography{2017_cvpr_pedestrian_detection}

\begin{thebibliography}{10}\itemsep=-1pt

\bibitem{Angelova2015Bmvc}
A.~Angelova, A.~Krizhevsky, V.~Vanhoucke, A.~Ogale, and D.~Ferguson.
\newblock Real-time pedestrian detection with deep network cascades.
\newblock In {\em BMVC}, 2015.

\bibitem{Benenson2014Eccvw}
R.~Benenson, M.~Omran, J.~Hosang, , and B.~Schiele.
\newblock Ten years of pedestrian detection, what have we learned?
\newblock In {\em ECCV, CVRSUAD workshop}, 2014.

\bibitem{cai16mscnn}
Z.~Cai, Q.~Fan, R.~Feris, and N.~Vasconcelos.
\newblock A unified multi-scale deep convolutional neural network for fast
  object detection.
\newblock In {\em ECCV}, 2016.

\bibitem{cai2015iccv}
Z.~Cai, M.~Saberian, and N.~Vasconcelos.
\newblock Learning complexity-aware cascades for deep pedestrian detection.
\newblock In {\em ICCV}, 2015.

\bibitem{Cordts2016Cityscapes}
M.~Cordts, M.~Omran, S.~Ramos, T.~Rehfeld, M.~Enzweiler, R.~Benenson,
  U.~Franke, S.~Roth, and B.~Schiele.
\newblock The cityscapes dataset for semantic urban scene understanding.
\newblock In {\em Proc. of the IEEE Conference on Computer Vision and Pattern
  Recognition (CVPR)}, 2016.

\bibitem{CosteaCVPR16}
A.~D. Costea and S.~Nedevschi.
\newblock Semantic channels for fast pedestrian detection.
\newblock In {\em CVPR}, 2016.

\bibitem{Dalal2005Cvpr}
N.~Dalal and B.~Triggs.
\newblock Histograms of oriented gradients for human detection.
\newblock In {\em CVPR}, 2005.

\bibitem{Dollar2014Pami}
P.~Doll\'ar, R.~Appel, S.~Belongie, and P.~Perona.
\newblock Fast feature pyramids for object detection.
\newblock {\em PAMI}, 2014.

\bibitem{Dollar2012Pami}
P.~Doll\'ar, C.~Wojek, B.~Schiele, and P.~Perona.
\newblock Pedestrian detection: An evaluation of the state of the art.
\newblock {\em PAMI}, 2012.

\bibitem{Enzweiler2009PAMI}
M.~Enzweiler and D.~M. Gavrila.
\newblock Monocular pedestrian detection: Survey and experiments.
\newblock {\em PAMI}, 2009.

\bibitem{Ess2008Cvpr}
A.~Ess, B.~Leibe, K.~Schindler, and L.~{Van Gool}.
\newblock A mobile vision system for robust multi-person tracking.
\newblock In {\em CVPR}, 2008.

\bibitem{Geiger2012CVPR}
A.~Geiger, P.~Lenz, and R.~Urtasun.
\newblock Are we ready for autonomous driving? the kitti vision benchmark
  suite.
\newblock In {\em CVPR}, 2012.

\bibitem{FastRCNN}
R.~Girshick.
\newblock Fast r-cnn.
\newblock In {\em ICCV}, 2015.

\bibitem{Girshick2014Cvpr}
R.~Girshick, J.~Donahue, T.~Darrell, and J.~Malik.
\newblock Rich feature hierarchies for accurate object detection and semantic
  segmentation.
\newblock In {\em CVPR}, 2014.

\bibitem{He2016Cvpr}
K.~He, X.~Zhang, S.~Ren, and J.~Sun.
\newblock Deep residual learning for image recognition.
\newblock In {\em CVPR}, 2016.

\bibitem{Hosang2015Cvpr}
J.~Hosang, M.~Omran, R.~Benenson, and B.~Schiele.
\newblock Taking a deeper look at pedestrians.
\newblock In {\em CVPR}, 2015.

\bibitem{Hu2016arxiv}
Q.~Hu, P.~Wang, C.~Shen, A.~van~den Hengel, and F.~Porikli.
\newblock Pushing the limits of deep cnns for pedestrian detection.
\newblock {\em ArXiv}, 2016.

\bibitem{densebox}
L.~Huang, Y.~Yang, Y.~Deng, and Y.~Yu.
\newblock Densebox: Unifying landmark localization with end to end object
  detection.
\newblock {\em arXiv}, 2015.

\bibitem{adamICLR15}
D.~Kingma and J.~Ba.
\newblock Adam: A method for stochastic optimization.
\newblock In {\em ICLR}, 2015.

\bibitem{Krizhevsky2012Nips}
A.~Krizhevsky, I.~Sutskever, and G.~E. Hinton.
\newblock Imagenet classification with deep convolutional neural networks.
\newblock In {\em NIPS}, 2012.

\bibitem{SA-FastRCNN}
J.~Li, X.~Liang, S.~Shen, T.~Xu, and S.~Yan.
\newblock Scale-aware fast r-cnn for pedestrian detection.
\newblock {\em arXiv}, 2016.

\bibitem{Li2016Eccv}
K.~Li and J.~Malik.
\newblock Amodal instance segmentation.
\newblock In {\em ECCV}, 2016.

\bibitem{liu15ssd}
W.~Liu, D.~Anguelov, D.~Erhan, C.~Szegedy, S.~Reed, C.-Y. Fu, and A.~C. Berg.
\newblock {SSD}: Single shot multibox detector.
\newblock In {\em ECCV}, 2015.

\bibitem{FCN15cvpr}
J.~Long, E.~Shelhamer, and T.~Darrell.
\newblock Fully convolutional models for semantic segmentation.
\newblock In {\em CVPR}, 2015.

\bibitem{YOLO}
J.~Redmon, S.~Divvala, R.~Girshick, and A.~Farhadi.
\newblock You only look once: Unified, real-time object detection.
\newblock In {\em CVPR}, 2016.

\bibitem{renNIPS15fasterrcnn}
S.~Ren, K.~He, R.~Girshick, and J.~Sun.
\newblock Faster {R-CNN}: Towards real-time object detection with region
  proposal networks.
\newblock In {\em Advances in Neural Information Processing Systems ({NIPS})},
  2015.

\bibitem{tian2015iccv}
Y.~Tian, P.~Luo, X.~Wang, and X.~Tang.
\newblock Deep learning strong parts for pedestrian detection.
\newblock In {\em ICCV}, 2015.

\bibitem{Yang2015Cvpr}
Y.~Tian, P.~Luo, X.~Wang, and X.~Tang.
\newblock Pedestrian detection aided by deep learning semantic tasks.
\newblock In {\em CVPR}, 2015.

\bibitem{Wojek2009Cvpr}
C.~Wojek, S.~Walk, and B.~Schiele.
\newblock Multi-cue onboard pedestrian detection.
\newblock In {\em CVPR}, 2009.

\bibitem{Wolf2006Ijcv}
L.~Wolf and S.~M. Bileschi.
\newblock A critical view of context.
\newblock {\em IJCV}, 2006.

\bibitem{Zhang2016Eccv}
L.~Zhang, L.~Lin, X.~Liang, and K.~He.
\newblock Is faster r-cnn doing well for pedestrian detection?
\newblock In {\em ECCV}, 2016.

\bibitem{shanshan_cvpr16}
S.~Zhang, R.~Benenson, M.~Omran, J.~Hosang, and B.~Schiele.
\newblock How far are we from solving pedestrian detection?
\newblock In {\em CVPR}, 2016.

\bibitem{Zhang2015Cvpr}
S.~Zhang, R.~Benenson, and B.~Schiele.
\newblock Filtered channel features for pedestrian detection.
\newblock In {\em CVPR}, 2015.

\end{thebibliography}

\clearpage{}

\appendix

\part*{Supplementary material}

\section{\label{sec:Content}Content}

In this supplementary material, we will show some more illustrations,
discussions and experiments for the CityPersons dataset.
\begin{itemize}
\item Section \ref{sec:CityPersons-dataset} shows some examples of our
annotations.
\item Section \ref{sec:Analysis-of-CityPersons} provides some analysis
regarding the annotations, including height statistics (section \ref{subsec:Height-histogram}),
analysis experiments regarding annotation quality (section \ref{subsec:Quality}).
\end{itemize}

\section{\label{sec:CityPersons-dataset}CityPersons annotation examples}

\begin{figure*}
\begin{centering}
\subfloat[Aachen]{\begin{centering}
\includegraphics[width=1\columnwidth]{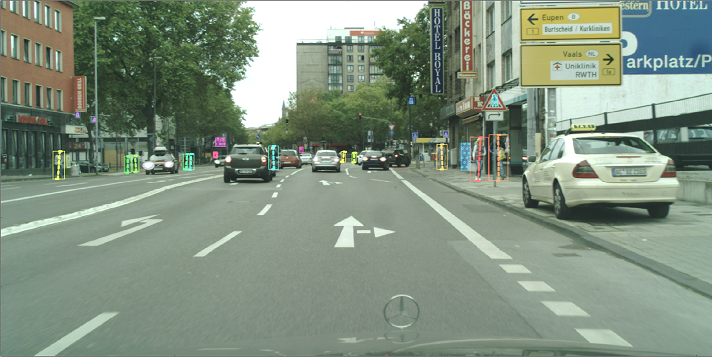}\ \includegraphics[width=1\columnwidth]{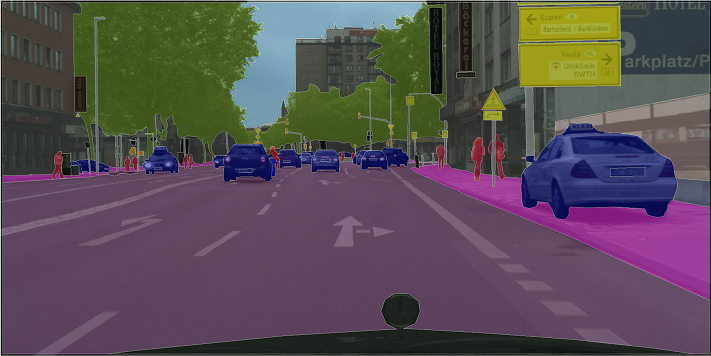}
\par\end{centering}
}
\par\end{centering}
\begin{centering}
\subfloat[Cologne]{\begin{centering}
\includegraphics[width=1\columnwidth]{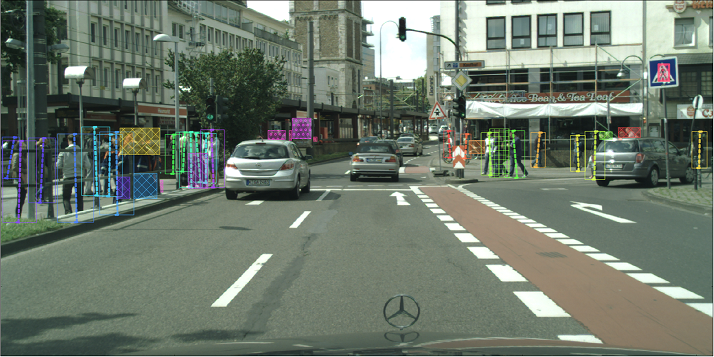}\ \includegraphics[width=1\columnwidth]{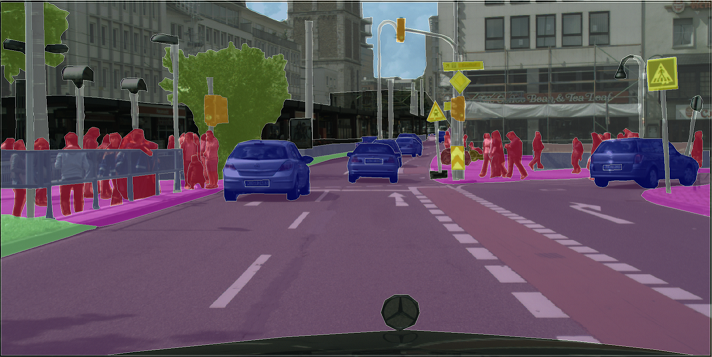}
\par\end{centering}
}
\par\end{centering}
\begin{centering}
\subfloat[Strasbourg]{\centering{}\includegraphics[width=1\columnwidth]{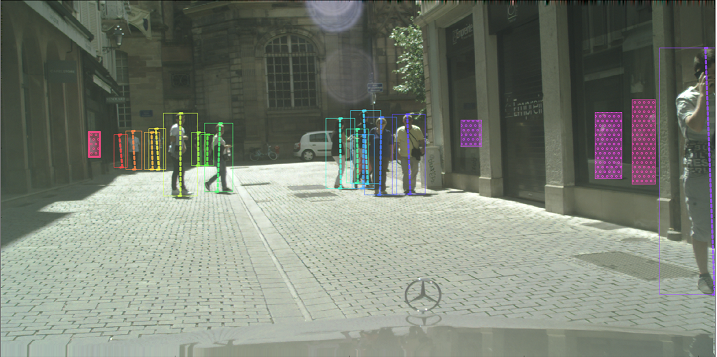}\ \includegraphics[width=1\columnwidth]{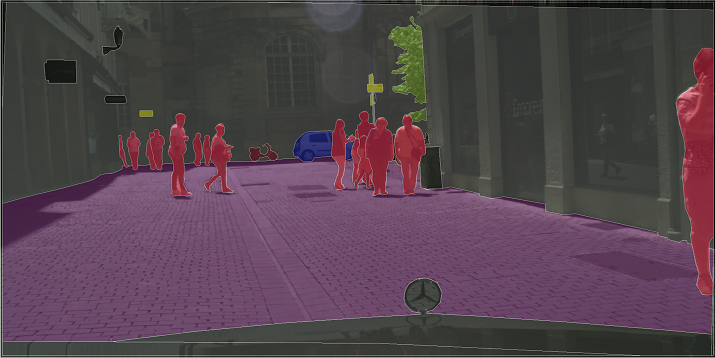}}
\par\end{centering}
\begin{centering}
\subfloat[Zurich]{\begin{centering}
\includegraphics[width=1\columnwidth]{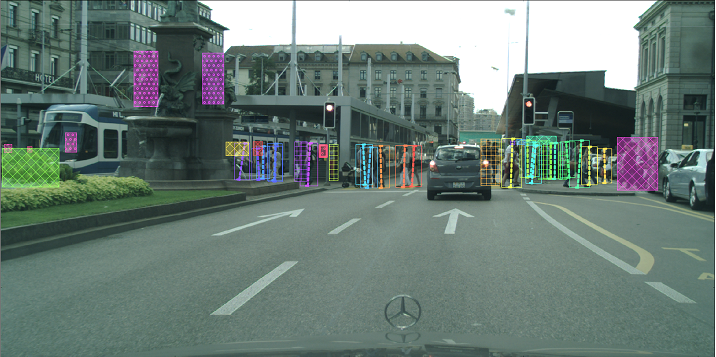}\ \includegraphics[width=1\columnwidth]{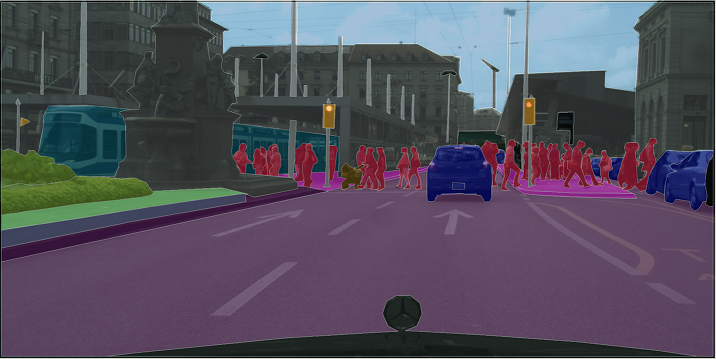}
\par\end{centering}
}
\par\end{centering}
\caption{\label{fig:Examples-of-annotations}Examples of annotations from different
cities. Left: our bounding box annotations; right: Cityscapes segmentation
annotations. For visualization, we use different masks for pedestrians/riders,
sitting persons, other persons, group of people, and ignore regions.}
\end{figure*}

In figure \ref{fig:Examples-of-annotations}, we show some examples
of our bounding box annotations and Cityscapes segmentation annotations
from different cities. We can see the diversity in terms of people's
appearance, clothing, and background objects.

\section{\label{sec:Analysis-of-CityPersons}Analysis of CityPersons annotations}

In this section, we provide some analysis regarding the height statistics
and quality of CityPersons annotations.

\subsection{\label{subsec:Height-histogram}Height statistics}

\begin{figure*}
\begin{centering}
\includegraphics[width=0.8\textwidth]{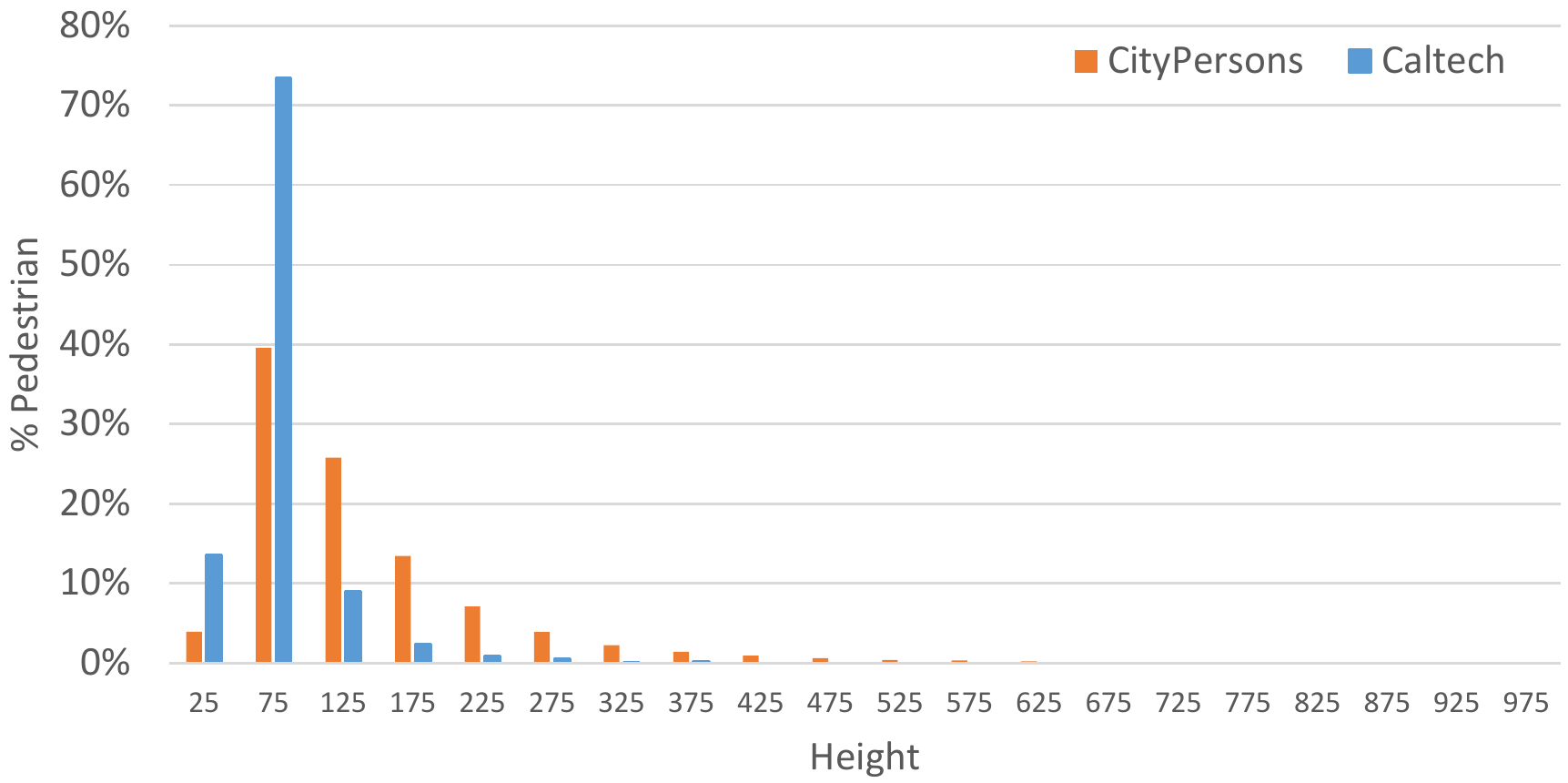}
\par\end{centering}
\caption{\label{fig:Height-histograms}Height distributions of CityPersons
and Caltech.}
\end{figure*}

In figure \ref{fig:Height-histograms}, we compare the height distribution
of CityPersons and Caltech. The CityPersons is more diverse than Caltech
in terms of scale:

(1) CityPersons covers a larger range of height, as it consists of
larger images. 

(2) More than 70\% of Caltech pedestrians fall in one single bin {[}50,100{]},
while CityPersons are more evenly distributed in different scale ranges. 

\subsection{\label{subsec:Quality}Quality}

The segment for each person only reflects the visible part, while
losing information of the aligned full body. In \cite{shanshan_cvpr16},
it is shown that better alignment of training annotations improve
the detection quality a lot. Therefore, in this paper we aim to provide
high quality well aligned annotations for each pedestrian. On the
other hand, as shown in the second section of the main paper, properly
handling ignore regions also affects the results, so we also make
efforts to label ignore regions over all images. 

In table \ref{tab:effects-quality}, we show that our high quality
annotations improve the performance by \textasciitilde{}7 pp, among
which \textasciitilde{}6 pp is gained from better alignment, and another
\textasciitilde{} 1pp from ignore regions handling.

Another argument for our aligned bounding box annotations is the comparison
of performance on an external benchmark (Caltech) using two types
of training annotations. From table \ref{tab:seg-bb-cross-eval},
we can see the model trained with segment bounding boxes fails not
only on CityPersons, but also on Caltech. The reason is other benchmarks,
e.g. Caltech, also provide aligned bounding box annotations. Therefore,
using our annotations helps to train a better generalizable model
over multiple benchmarks.

\begin{table}
\centering{}%
\begin{tabular}{c|cc}
Annotation aspect & $\mbox{MR}$  & $\Delta\mbox{MR}$\tabularnewline
\hline 
\hline 
segment bounding boxes  & 22.54 & -\tabularnewline
+ ignore regions & 21.31 & + 1.23\tabularnewline
+ better boxes & 15.14 & + 6.17\tabularnewline
\hline 
our annotations & 15.14 & + 7.40\tabularnewline
\end{tabular}\caption{\label{tab:effects-quality}The effects on performance of using high
quality training annotations. CityPersons validation set evaluation.
Training with our aligned bounding box annotations and ignore region
annotations gives better performance than training with segment bounding
box annotations.}
\end{table}

\begin{table}
\centering{}%
\begin{tabular}{cc|c|c}
 & Train anno. & Seg. & Aligned \tabularnewline
Test set &  & bounding box & bounding box\tabularnewline
\hline 
\hline 
\multicolumn{2}{c|}{Caltech} & 37.5 & 26.9\tabularnewline
\multicolumn{2}{c|}{CityPersons} & 22.5 & 15.1\tabularnewline
\end{tabular}\caption{\label{tab:seg-bb-cross-eval}Comparison of performance using two
types of training annotations. Numbers are MR on CityPersons validation
set; and $\mbox{MR}^{O}$ on Caltech test set. Using our aligned bounding
box for training obtains better quality on both Caltech and CityPersons.}
\end{table}

\clearpage{}
\end{document}